\documentclass[letterpaper, 10 pt, conference]{ieeeconf}  
\usepackage[margin=0.792in]{geometry}
\usepackage{cite}
\usepackage{amsmath,amssymb,amsfonts}
\usepackage{algorithmic}
\usepackage{caption} 
\usepackage{booktabs} 
\usepackage{graphicx}
\usepackage{textcomp}
\usepackage{xcolor}
\usepackage{url}
\usepackage{subcaption}
\usepackage{cleveref}
\usepackage{tabularx}   
\usepackage{multirow}   
\usepackage[table]{xcolor} 
\definecolor{bestcolor}{rgb}{0.9, 0.9, 1} 

\IEEEoverridecommandlockouts                              

\title{\LARGE \bf
Learning-based Cooperative Robotic Paper Wrapping: A Unified Control Policy with Residual Force Control 
}

\author{Rewida Ali$^{1}$ , Cristian C. Beltran-Hernandez$^{2}$, Weiwei Wan$^{1}$ , and Kensuke Harada$^{1,3}$ 
\thanks{$^{1}$Department of Systems Innovation, Graduate School of Engineering Science, Osaka University, Osaka 560-0043, Japan. 
        {\tt\small rewidaali@hlab.sys.es.osaka-u.ac.jp, wan@sys.es.osaka-y.ac.jp.}}%
\thanks{$^{2}$OMRON SINIC X Corporation, Tokyo, Japan
        {\tt\small cristian.beltran@sinicx.com.}}%
\thanks{$^{3}$The National Institute of Advanced Industrial Science and Technology (AIST), Tokyo 135-0064, Japan
        {\tt\small harada@sys.es.osaka-y.ac.jp.}}
}

\begin{document}

\maketitle
\thispagestyle{empty}
\pagestyle{empty}

\begin{abstract}
Human–robot cooperation is essential in environments such as warehouses and retail stores, where workers frequently handle deformable objects like paper, bags, and fabrics. Coordinating robotic actions with human assistance remains difficult due to the unpredictable dynamics of deformable materials and the need for adaptive force control. To explore this challenge, we focus on the task of gift wrapping, which exemplifies a long-horizon manipulation problem involving precise folding, controlled creasing, and secure fixation of paper. Success is achieved when the robot completes the sequence to produce a neatly wrapped package with clean folds and no tears.

We propose a learning-based framework that integrates a high-level task planner powered by a large language model (LLM) with a low-level hybrid imitation learning (IL) and reinforcement learning (RL) policy. At its core is a Sub-task Aware Robotic Transformer (START) that learns a unified policy from human demonstrations. The key novelty lies in capturing long-range temporal dependencies across the full wrapping sequence within a single model. Unlike vanilla Action Chunking with Transformer (ACT), typically applied to short tasks, our method introduces sub-task IDs that provide explicit temporal grounding. This enables robust performance across the entire wrapping process and supports flexible execution, as the policy learns sub-goals rather than merely replicating motion sequences.

Our framework achieves a 97\% success rate on real-world wrapping tasks. We show that the unified transformer-based policy reduces the need for specialized models, allows controlled human supervision, and effectively bridges high-level intent with the fine-grained force control required for deformable object manipulation.
\end{abstract}

\section{INTRODUCTION}

\begin{figure}[t]
      \centering
      \includegraphics[width=\columnwidth]{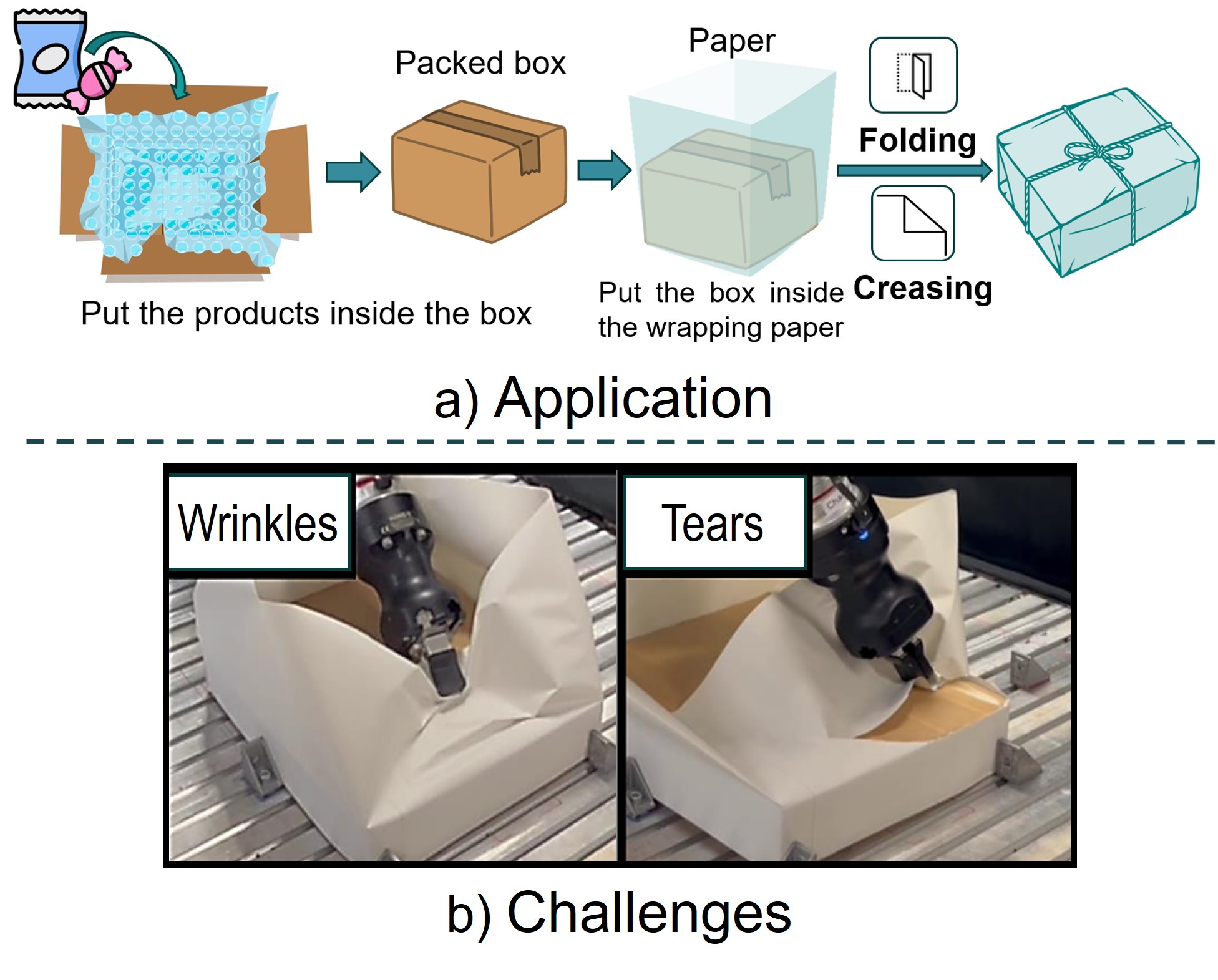}
      \caption{\textbf{System Overview:} (a) Application Context: Wrapping as the final stage in an automated packaging (b) Key Challenges: Wrinkles, and tears.}
      \label{fig:1}
   \end{figure}

Robotic manipulation of deformable objects such as paper is critical for applications ranging from logistics to assistive robotics. Consider a typical e-commerce fulfillment pipeline: products are autonomously retrieved from shelves \cite{chen2022category} and packed using deterministic algorithms, yet the final gift-wrapping step—a multi-stage process involving folding, tucking, and adhesive application—still relies heavily on manual labor (Fig.\ref{fig:1}(a)). While humans perform such tasks intuitively, robots struggle to integrate high-level sequencing with low-level control, limiting real-world deployment \cite{li2025large}. Closing this automation gap requires a unified framework that combines robust perception, adaptive planning, and precise control for deformable materials. Unlike highly compliant fabrics, paper exhibits significant bending stiffness that necessitates a two-stage deformation process: initial folding to achieve the target geometry, followed by precise creasing to permanently set the shape through localized plastic deformation (Fig.\ref{fig:1}a).

Paper wrapping presents unique physical challenges (Fig.~\ref{fig:1}b): (1)  unpredictable wrinkle formation demands real-time visual correction; (2) low tear resistance requires precise force modulation to prevent damage; and (3) bimanual coordination is often essential for complex folds. Current approaches struggle with these demands \cite{haslach2009time}. Pure RL methods \cite{9732695} suffer from sample inefficiency, while goal-conditioned policies \cite{mo2022foldsformer} fail when future states are uncertain, such as during non-linear deformation of the material. While demonstration-based methods can address these issues, they require large datasets that are particularly difficult to collect for this task due to the paper's non-linear behavior and visual ambiguities caused by its color.

Conventional demonstration-based methods also struggle with capturing fine-grained force profiles and often rely on hardware-specific tele-operation setups \cite{fu2024mobile,wu2024gello}. To overcome this, we adopt a VR-based demonstration system \cite{kamijo2024learning} that enables natural bimanual control, precise 6D pose tracking, variable gripper control, and high-frequency sampling (40 Hz). Our setup also integrates programmable haptic feedback, mapping contact forces to haptic vibrations, and ensures safety through compliance control. This allows us to capture human expertise in wrinkle mitigation and tear prevention, which we transfer through the IL model and refine with a residual RL module.

To address the gift wrapping task, a complex manipulation challenge, we develop a hybrid learning framework that synergizes human expertise with data-driven policy learning. Our approach is structured hierarchically to tackle the long-horizon nature of the task.

First, to manage the complexity and variability of the wrapping sequence, we decompose the task into a number of semantically meaningful sub-tasks guided by human demonstrations. A fine-tuned large language model (LLM) acts as a high-level planner, translating natural language commands from a human assistant into executable robot actions. This enables flexible, language-guided synchronization for human–robot collaboration, allowing the task sequence to be dynamically adapted based on verbal instructions. It also allows human engagement when required, by inserting appropriate waiting times into the robot’s execution.

For low-level action generation, we introduce the Sub-Task Aware Robotic Transformer (START), a novel improvement of the ACT model \cite{zhao2023learning}. This model learns a unified policy across all sub-tasks by conditioning its predictions on a learnable sub-task ID marker. This architectural innovation allows a single transformer to capture temporal context within the long-horizon task, thereby enabling robust transitions between distinct phases of manipulation.

Finally, to ensure precise and safe execution in the real world, the actions from START are refined by a constrained residual policy. This module, trained with reinforcement learning, provides real-time admittance-based corrections. It adjusts the robot's end-effector positions and the low-level controller parameters, allowing the robot to adapt compliantly to unexpected contact and material deformation.

Bringing these components together, our framework operates in Cartesian space, conditioned on 6D force/torque (F/T) data and the current sub-task ID. In summary, our key contributions are:

\begin{enumerate}
\item \textbf{LLM-based Task Planner}: A language-guided planner that synchronizes human–robot collaboration by translating natural language assistance requests into executable robot commands, enabling flexible task sequencing.
\item \textbf{Sub-Task Aware Robotic Transformer (START)}: A transformer-based architecture that incorporates learnable sub-task IDs to acquire a unified, temporally-aware policy for long-horizon object manipulation.
\item \textbf{Constrained Residual Policy}: An interaction-based RL module that ensures safe, precise actions by applying bounded Cartesian corrections ($\Delta \mathbf{x}$) and optimizing admittance parameters ($\mathbf{K}$) in real time.
\end{enumerate}
\section{Related Work} \label{Related work}

\subsection{Manipulation of Deformable Objects}
Research on deformable object manipulation mainly follows two approaches: model-based and learning-based. Model-based methods, such as finite element \cite{hu2018three,kita2004deformable} and particle-based simulations \cite{chen2024differentiable}, offer accurate deformation prediction but require precise material parameters, struggle with real-time performance, and fail to capture stiffness properties critical for paper manipulation. Learning-based methods instead map observations to actions. While early works \cite{wang2022learning,mo2022foldsformer} showed task-specific success, they lacked generalization. Recent advances introduce language conditioning \cite{deng2024learning}, semantic keypoints \cite{sundaresan2023kite}, or hybrid IL/RL frameworks \cite{hanai2025robotic}, yet they remain limited to short horizons or rely on task-specific representations, necessitating separate models for different actions. UniFolding \cite{xue2023unifolding} predicts actions from point clouds but suffers from discretization errors and poor generalization to paper tasks.

\textbf{In this paper,} we address these challenges of generalization and multi-step operations by learning a unified set of admittance-based action primitives through sub-task decomposition. Our approach is not only easier to learn but also enables the reuse of skills across the entire long-horizon task.
   \begin{figure*}[t]
      \centering
      \includegraphics[width=0.95\textwidth]{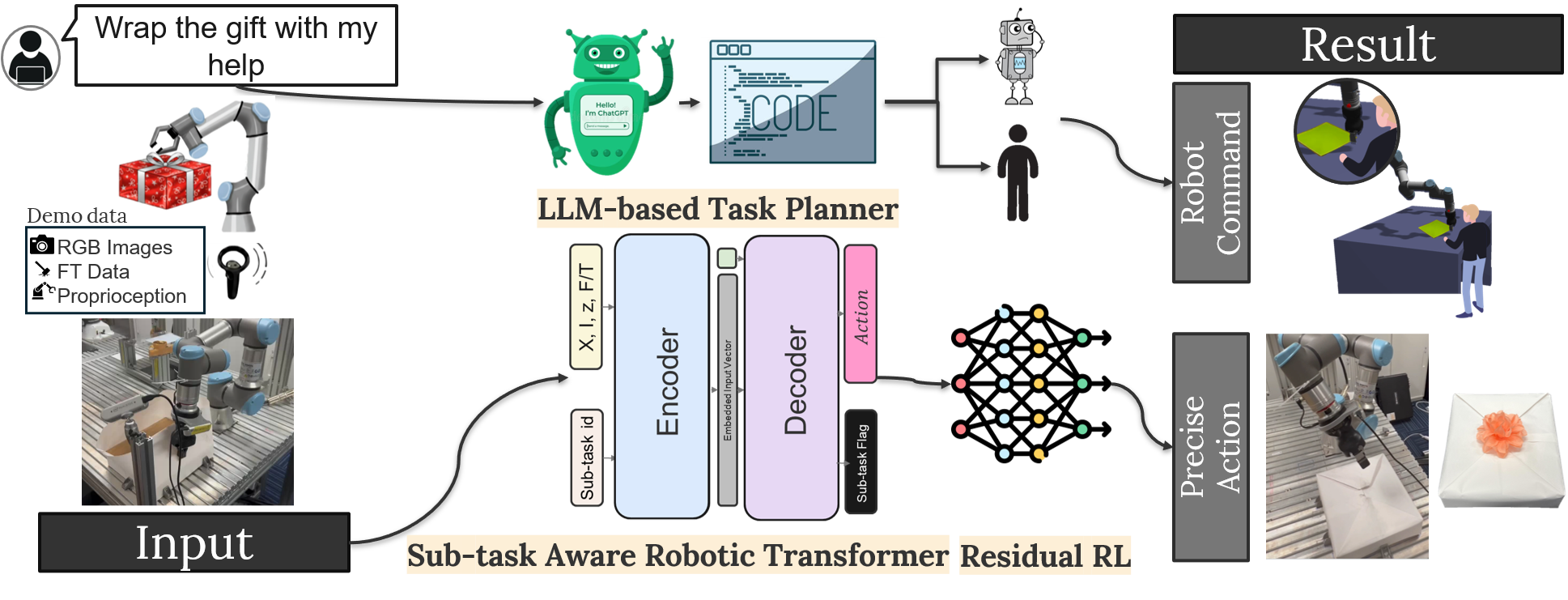}
      \caption{\textbf{Overview of the proposed framework:} the system takes as input a natural language task description, multi-view RGB observations, robot proprioception, and force/torque data. START model predicts high-level actions using a unified policy. In parallel, a language-based task planner synchronizes collaboration with the human partner, while a residual RL module learns admittance control parameters for the compliant execution of precise action.}
      \label{fig:2}
   \end{figure*}
\subsection{Learning Long-horizon Tasks}

Recent advances in machine learning have driven progress in data-driven deformable object manipulation (DOM). Early methods focused on direct imitation learning \cite{hanai2025robotic,lee2015learning,mitrano2022focused}, while recent work emphasizes latent representations that capture physical properties \cite{ganapathi2021learning,zhao2023learning}. Approaches such as Diffusion Policy \cite{chi2023diffusion} achieve strong results across diverse tasks but still face generalization challenges due to limited real-world demonstration data.

For long-horizon imitation learning, two main strategies have emerged: one-shot imitation, which learns entire tasks from a single demonstration via videos or trajectories \cite{chen2025vidbot,huang2019neural}, and hierarchical imitation, which decomposes tasks into high-level sub-goals and low-level controllers \cite{lynch2020learning,mandlekar2020iris}. The latter improves scalability by reducing planning frequency and enabling more robust execution over extended horizons.

Although effective, both approaches face limitations. One-shot methods require careful objective tuning to achieve robust performance, while hierarchical methods often fail to fully exploit the compositional structure of demonstrations, thereby limiting their generalization to novel task variations.

\textbf{Our approach} extends hierarchical imitation learning by unifying task decomposition and execution into a single transformer-based policy (START), while addressing prior limitations through: (1) learnable sub-task representations that explicitly model phase transitions in demonstrations, and (2) closed-loop action refinement via admittance-based RL. This hybrid paradigm combines the sample efficiency of imitation learning with the adaptability of hierarchical control.

\section{Methodology} \label{Method}

This work introduces a unified learning framework for long-horizon, fine-motor precision tasks, with a focus on the paper wrapping task. Our system integrates imitation learning, language-based task planning, and hybrid control to enable both autonomous and collaborative wrapping in unstructured environments. The overall architecture comprises four key components: (1) a language-based task planner for human–robot coordination, (2) A sub-task guided imitation learning model, and (3) a residual reinforcement learning controller for fine-grained force modulation. An overview of the system is shown in Fig. \ref{fig:2}.

\subsection{Problem Formulation}

We address the problem of learning an end-to-end policy that enables a robot to perform a complete paper wrapping task composed of multiple sequential manipulation subtasks.
Given a multi-modal observation history:
\begin{equation}
o_t = {I_t^{(n)}, S_t, f_t, sub_t}
\end{equation}
where $I_t^{(n)} \in \mathbb{R}^{H \times W \times 3}$ are RGB images from different viewpoints, $S_t$ is the robot state including end-effector position, orientation, and gripper aperture, $f_t \in \mathbb{R}^6$ is the force–torque (F/T) sensor reading, and $sub_t$ is the sub-task context embedding, our objective is to produce a control action:
\begin{equation}
a_t = {p_t, R_t^{(6)}, g_t}
\end{equation}
where $p_t \in \mathbb{R}^3$ is the end-effector Cartesian translation, $R_t^{(6)} \in \mathbb{R}^6$ is the orthogonal 6D rotation representation, and $g_t \in \mathbb{R}$ is the gripper aperture command.

\subsection{LLM-based Task Planner} \label{sec:planner}

In deformable object manipulation (DOM) tasks—such as paper wrapping—the execution steps can vary significantly depending on object size, paper stiffness, and the degree of human involvement. For instance, a human operator may only partially assist (e.g., holding a corner) or request the robot to perform the entire task autonomously. Designing fixed scripts to handle all such variations quickly becomes infeasible. Instead, natural language offers an intuitive interface for specifying diverse objectives in a flexible, high-level manner.

To address this, we propose an LLM-based task planning module with three core objectives: (1) generating sub-task identifiers (IDs) used by the learning model to produce actions, (2) decomposing task descriptions into structured steps, and (3) transforming these steps into executable robot commands. The module receives a natural language instruction from the user—such as \texttt{wrap the box from the left side}—and processes it in two parallel pathways. First, a fine-tuned GPT-4o \cite{hurst2024gpt} generates a complete sequence of sub-task IDs in a single pass. These IDs provide flexibility, allowing the robot to adapt the wrapping order (e.g., starting from the left, right, or a transient state). Second, a fine-tuned GPT-3.5~ \cite{openai2022chatgpt} decomposes the same instruction into low-level steps that can be mapped to robot primitives.
   \begin{figure}[t]
      \centering
      \includegraphics[width=0.5\textwidth]{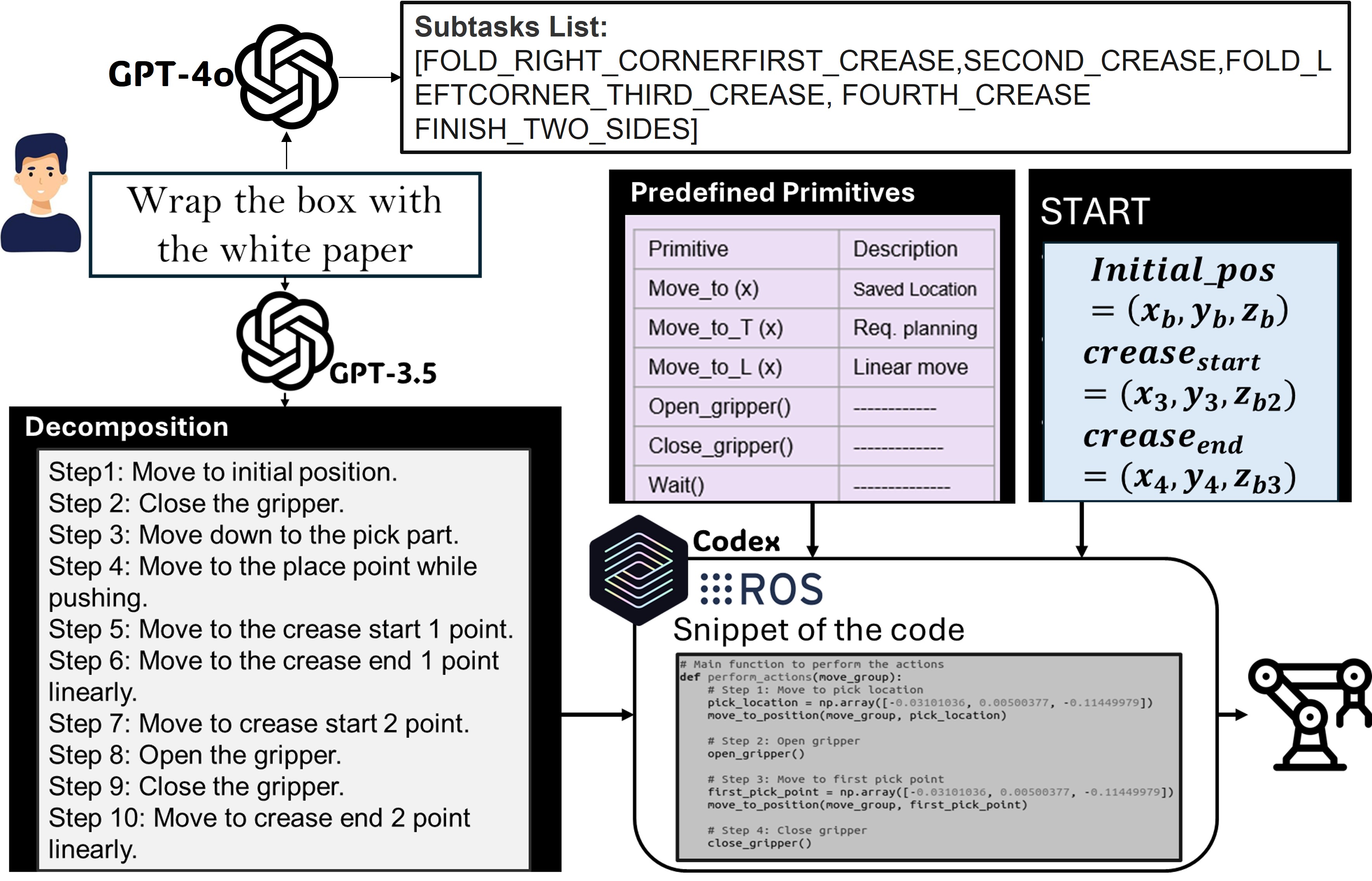}
      \caption{Task planner framework based on LLMs: it consists of two steps, first dividing the task description into steps using GPT. Then, by using Codex to generate executable robot commands by combining the steps with predefined primitives and the coordinates generated from the transformer-based learning model. In parallel, it also generates sub-task IDs.}
      \label{fig:4}
   \end{figure}
For training, GPT-4o was fine-tuned on 10 curated examples, each mapping a high-level task description to a sequence of sub-task IDs corresponding to robot actions. GPT-3.5 was fine-tuned on 20 examples to produce detailed, low-level steps. For instance, given the instruction \texttt{Wrap the box with the white paper}, the GPT-4o branch outputs the sequence of sub-task IDs for conditioning the START model, while the GPT-3.5 branch generates structured action steps (see Fig.~\ref{fig:4}). These steps are then translated into executable robot code using Codex \cite{chen2021evaluating}, which maps them to a predefined set of 12 primitives (e.g., \texttt{move\_to}, \texttt{open\_gripper}, \texttt{wait\_for\_human}). Finally, data transfer within the module is handled through the Robot Operating System (ROS), enabling robustness and portability to other ROS-based platforms.
\subsection{Sub-task Aware Robotic Transformer Model}

Our model, START, extends temporal transformers for long-horizon control by introducing a sub-task identifier mechanism. Tasks are decomposed into sub-tasks, each labeled with a text-based ID, allowing a single unified policy to condition action generation on the current sub-task. Given multimodal inputs (RGB, proprioception, and force), START outputs structured robot actions, enabling the robust execution of complex human–robot cooperation tasks involving deformable materials.

\begin{figure}[t]
      \centering
      \includegraphics[scale=0.32]{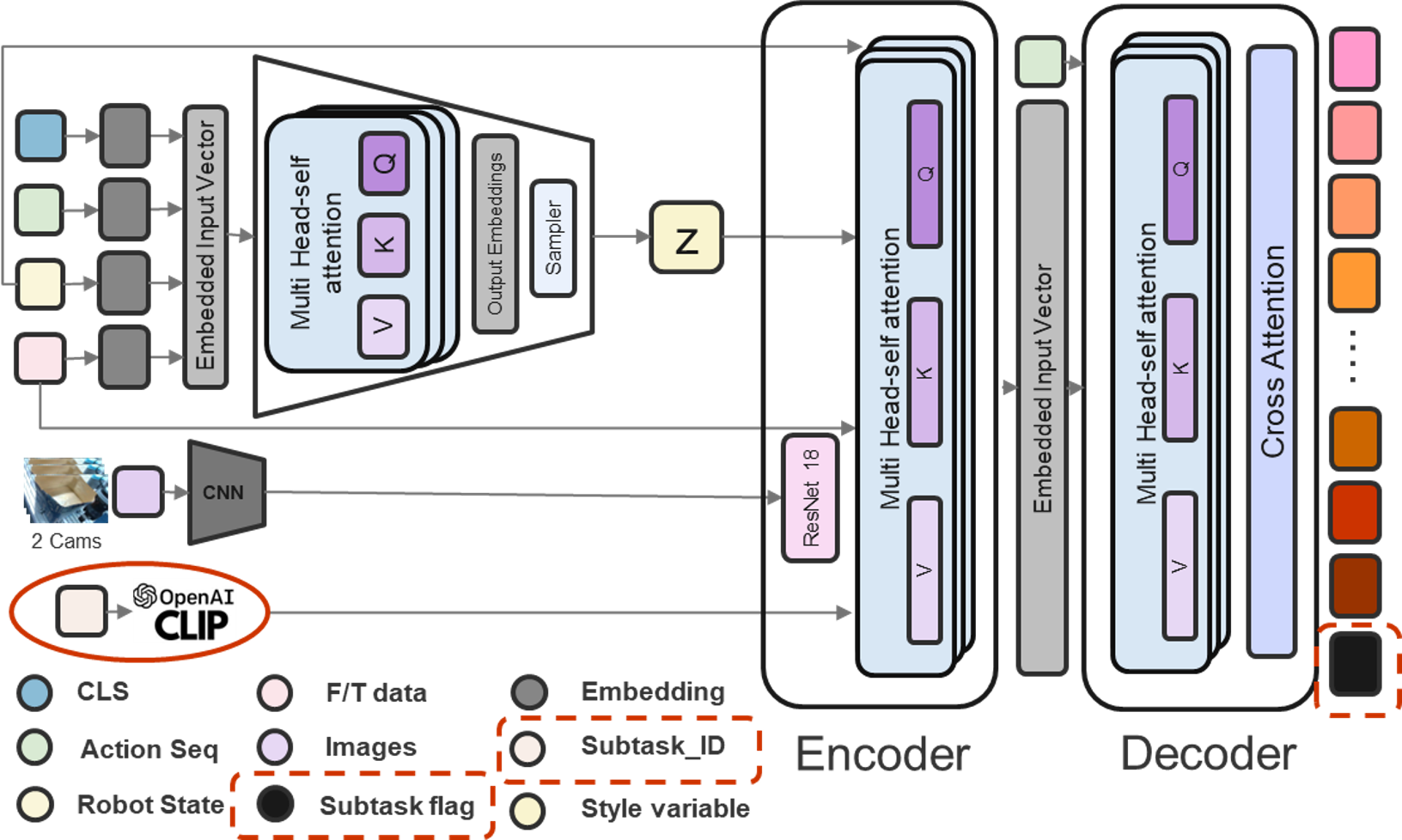}
      \caption{Proposed framework for learning an IL task in a real robot using a modified task-aware START model.}
      \label{fig:3}
   \end{figure}

In our implementation, illustrated in Fig.~\ref{fig:3}, the model operates as follows:
{\footnotesize
\begin{itemize}
    \item \textbf{Inputs:} 
    \begin{enumerate}
        \item Visual features $\phi(I_t^{(v)})$ via ResNet-18,
        \item Robot state $S_t$,
        \item Force/torque readings $f_t$,
        \item Sub-task embeddings $s_t = \text{CLIP}(d_t)$.
    \end{enumerate}
    \item \textbf{Outputs:} 
    \begin{enumerate}
        \item Cartesian action $\{p_t, R_t^{(6)}, g_t\}$,
        \item Transition flag $\delta_t \in \{0,1\}$.
    \end{enumerate}
\end{itemize}
}

Sub-task IDs generated by the planner (Section~\ref{sec:planner}) support long-horizon execution by structuring demonstrations into sub-tasks $\tau_i$. Each sub-task is paired with a CLIP-based language embedding ~\cite{thengane2022clip}, injected into the transformer encoder to provide context-dependent conditioning. This enables the model to specialize motion patterns (e.g., \texttt{pick} vs. \texttt{crease}) while maintaining a single unified policy. For new tasks, the sequence is divided into sub-tasks, with IDs defined manually or via LLM, and automatically assigned by the planner from human task descriptions. In addition to predicting actions, the model outputs a transition flag that enables implicit switching between motion modes.




\subsection{Residual RL for Admittance Parameter Adaptation}

In deformable object tasks, such as paper wrapping, purely position-based control often fails to adapt to subtle variations in contact dynamics. To overcome this, we integrate a residual admittance control module that enables compliant execution and adaptation to local forces. At first glance, one might ask: \textbf{Why is force control essential in paper wrapping?} First, it supports generalization across papers with different stiffness, where a single position-based policy would fail \cite{choi2024learning}. Second, force control is necessary in dynamic scenarios during folding, where the natural behavior of the paper results in inconsistent creasing conditions \cite{elbrechter2012folding}. In such cases, due to local variations in tension and resistance, the required force along the creasing edge may not be constant. This further underscores the importance of modulating force through compliant control or by refining the action using an admittance-based adjustment mechanism.

Therefore, this module enables the robot to learn corrective adjustments conditioned on contact dynamics. We adopt Soft Actor-Critic (SAC) for this purpose due to its robustness in continuous action spaces and its sample efficiency, which makes it suitable for learning in the real world. The RL agent receives, as input at each timestep, a state vector composed of key contact and tracking features:
\begin{equation}
s_t = [\Delta P, \dot{P}_t, f_t, f^D_{t-1}],
\end{equation}
where \( \Delta P = P_t - P^D_{\text{START}} \) represents the position error with respect to the START-predicted desired waypoint, \( \dot{P}_t \) is the current tool velocity, \( f_t \) is the measured contact force, and \( f^D_{t-1} \) denotes the desired contact force from the previous timestep.

The output of the RL policy is a residual action vector defined as:
\begin{equation}
a_t = [\Delta P, K_p, K_d],
\end{equation}
where \( K_p \) and \( K_d \) are adaptive proportional and derivative gains learned by the policy.

The updated admittance command is computed as:
\begin{equation}
F_{\text{cmd}} = K_p (P^D - P_t) + K_d (\dot{P}^D - \dot{P}_t),
\end{equation}
where \( P^D = P^D_{\text{START}} + \Delta P \) when residual pose correction is enabled.

To ensure stable and effective learning, we design a reward function that encourages accurate trajectory following, smooth force regulation, and penalizes instability:
\begin{equation}
r_t = -\alpha \| P^D - P_t \|^2 - \beta \| f_t - f^D \|^2 - \gamma \| \Delta f \|^2
\end{equation}
with \( \alpha > \beta > \gamma \), thereby prioritizing spatial accuracy while maintaining compliant and safe contact behavior.

   \begin{figure}[t]
      \centering
      \includegraphics[width = 0.5\textwidth]{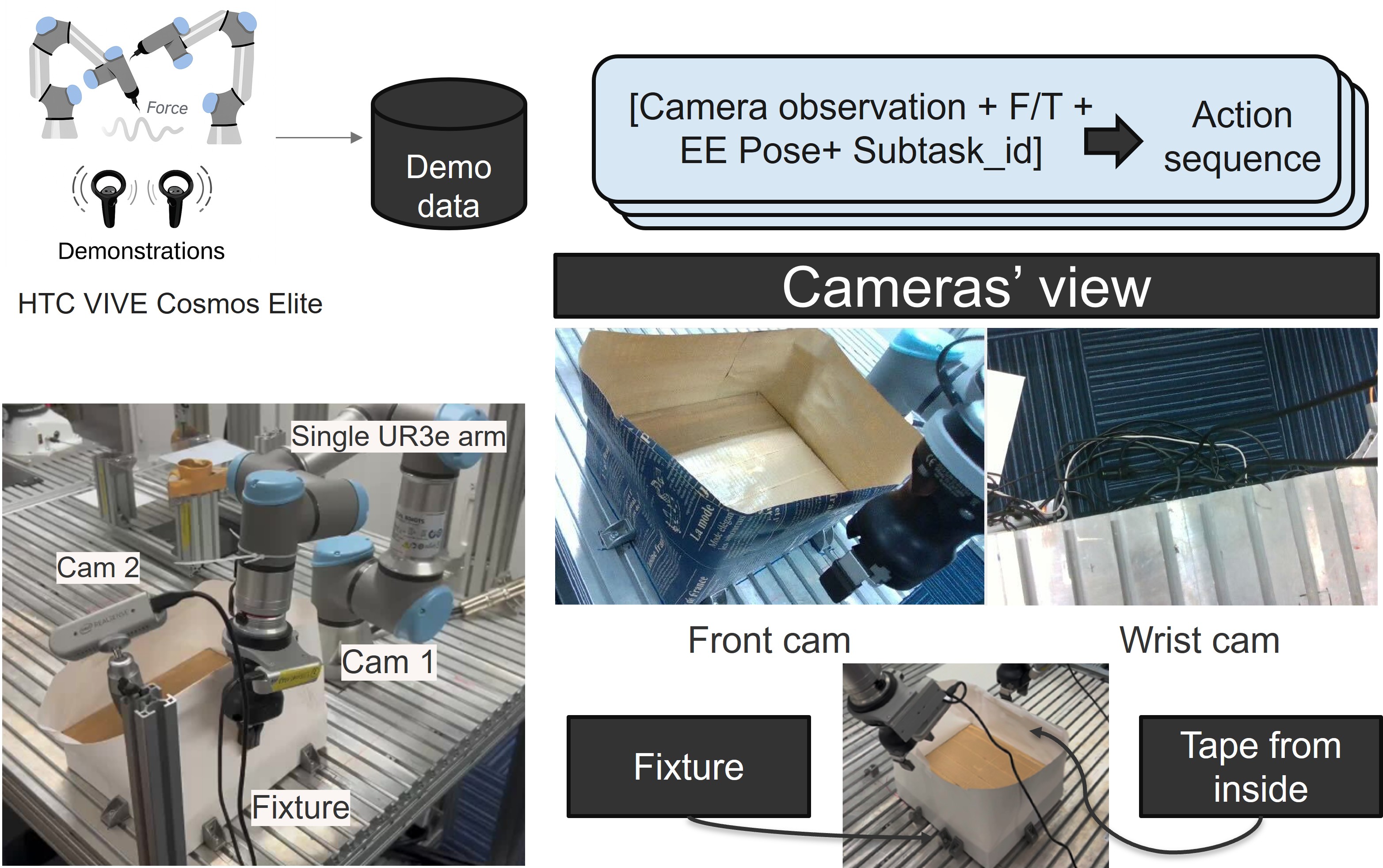}
      \caption{The experimental hardware setup featuring a UR3e robotic arm with a Robotiq Hand-E gripper. Two Intel RealSense D435 cameras provide multi-view RGB perception.}
      \label{fig:5}
   \end{figure}
\section{Experiments} \label{experiments}

To evaluate our proposed framework, we measured overall task success rates and assessed the quality of folded or wrapped paper. We then ablated each module to determine its individual performance contribution.

\subsection{Hardware Setup}

We validated our framework on a UR3e robotic arm with a Robotiq Hand-E gripper and force–torque sensor. The UR3e was selected for its industrial reliability and ROS compatibility. While UR robots are not torque-controlled and thus only support admittance control, our framework adapts this limitation to enable compliant manipulation tasks such as paper folding and wrapping.  

For visual perception, two Intel RealSense D435 RGB-D cameras were positioned at different viewpoints: one mounted on the wrist and the other placed diagonally. This multi-view setup enabled the system to capture the scene from complementary perspectives, as illustrated in Fig.~\ref{fig:5}.

The gift wrapping task used a standard industrial box (275 × 240 × 50 mm) and white printer paper (160 GSM, 0.20 mm thick, stiffness 6–12 mN·m, dimensions 180 × 1030 mm), representative of lightweight packaging. As shown in Fig.\ref{fig:5}, the box was laterally supported to prevent displacement without rigid fixation, ensuring realistic interaction. Due to the single-arm setup, small adhesive strips were applied to secure paper edges during folding; future dual-arm extensions will allow bimanual wrapping without adhesives. 
\subsection{Data Collection and Tasks}

\begin{figure}[t]
      \centering
      \includegraphics[width = 0.5\textwidth]{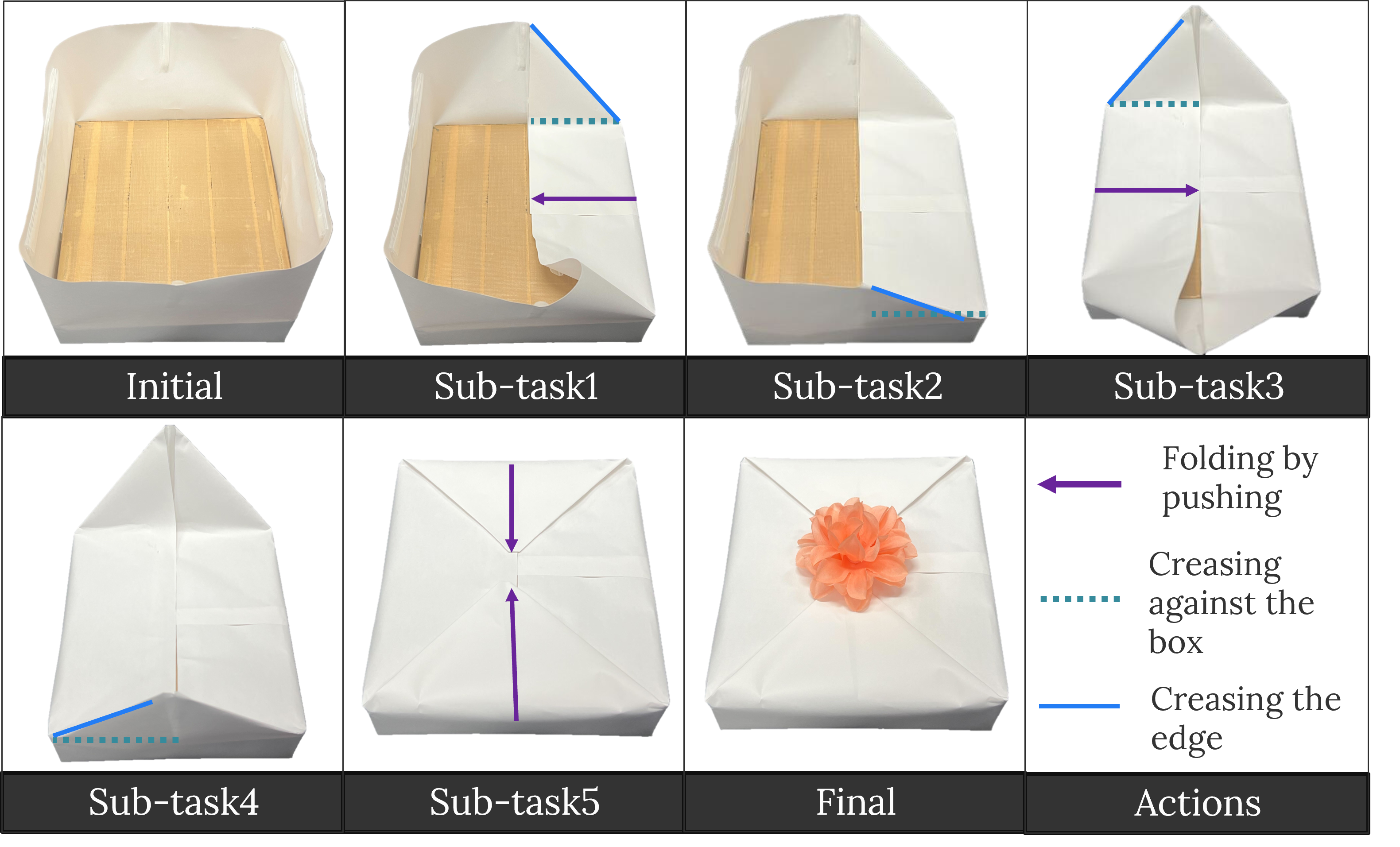}
      \caption{The five reusable subtasks and three core actions defining the paper wrapping policy[folding by pushing, creasing against the box, and creasing the edge].}
      \label{fig:6}
   \end{figure}

A structured dataset of expert trajectories was collected for the paper wrapping task, segmented into five reusable subtasks based on human wrapping behavior (Fig.~\ref{fig:6}): 
(1) push and crease one side and its edge, 
(2) crease the opposite side and edge, 
(3) repeat (1) on the opposite half, 
(4) repeat (2) on the same half, and 
(5) close both sides. Each sub-task included 50 demonstrations, totaling 250, all performed under a consistent initial setup. The main actions involved were: folding by pushing, creasing against the box, and creasing the edge. To achieve the best results on new tasks, we limit each sub-task to a maximum of 1,200 steps.

Demonstrations were recorded using a VR-based teleoperation system following the COMP-ACT framework \cite{kamijo2024learning}, enabling intuitive, real-time control of a virtual end-effector with position and force trajectories. Visual feedback via a VR headset and haptic cues allowed precise demonstration of long-horizon, contact-rich motions while capturing compliance and force adjustments with the deformable paper.

\section{Results} \label{results}

\subsection{Training and Robustness Evaluation}

The model was trained using the LeRobot framework \cite{cadene2024lerobot}, which supports the multi-modal inputs required for our task. Training was conducted for 5,000 epochs with a batch size of 10 on an NVIDIA RTX 4070 GPU, taking approximately 20 hours. Key hyperparameters included a cosine learning rate scheduler with a warm-up period and gradient clipping to ensure stable optimization.

\subsection{Evaluation of Task Completion}

The policy was evaluated on five sub-tasks, each executed 150 times. While the setup largely matched the demonstrations, variations were introduced, including camera orientation offsets of up to ±5$\sim$7° and different lighting conditions. Success was determined by human assessment of positioning, attachment, and absence of paper tears or major wrinkles. The results are shown in Table \ref{tab:success_rates}, with an overall average success rate of \textbf{97.3\%}.

\begin{table}[h]
\centering
\caption{Success rates for each sub-task}
\label{tab:success_rates}
\begin{tabular}{lcc}
\toprule
\textbf{Task} & \textbf{Successful Trials} & \textbf{Success Rate} \\
\midrule
Sub-task\_1 & 150/150 & 100\% \\
Sub-task\_2 & 150/150 & 100\% \\
Sub-task\_3 & 146/150 & 97.33\% \\
Sub-task\_4 & 144/150 & 96.33\% \\
Sub-task\_5 & 140/150 & 93.33\% \\
Whole-task &  146/150 & 97.33\% \\
\bottomrule
\end{tabular}
\end{table}

\begin{table}[b]
\centering
\caption{Paper integrity scores during wrapping. Lower T, W, and E indicate better quality, and higher PI indicates better quality.}
\label{tab:paper_integrity_results}
\rowcolors{2}{gray!10}{white}
\begin{tabular}{l>{\centering\arraybackslash}p{1.4cm}>{\centering\arraybackslash}p{0.9cm}>{\centering\arraybackslash}p{1.3cm}>{\centering\arraybackslash}p{1.0cm}}
\hline
\textbf{Evaluation Stage} & \textbf{Tears (T) $\downarrow$} & \textbf{Wrinkles (W) $\downarrow$} & \textbf{Edge (E) $\downarrow$} & \textbf{PI $\uparrow$} \\
\hline
After Sub-Task 1 & \cellcolor{green!30}0.00 & \cellcolor{red!20}0.22 & \cellcolor{red!20}0.22 & \cellcolor{red!20}0.890 \\
After Sub-Task 2 & \cellcolor{green!30}0.00 & \cellcolor{red!30}0.25 & \cellcolor{red!10}0.18 & \cellcolor{red!20}0.889 \\
After Sub-Task 3 & \cellcolor{green!30}0.00 & \cellcolor{red!10}0.20 & \cellcolor{yellow!20}0.16 & \cellcolor{yellow!20}0.908 \\
After Sub-Task 4 & \cellcolor{green!30}0.00 & \cellcolor{green!20}0.10 & \cellcolor{green!20}0.12 & \cellcolor{green!20}0.946 \\
After Sub-Task 5 & \cellcolor{red!10}0.02 & \cellcolor{green!30}0.06 & \cellcolor{green!40}0.05 & \cellcolor{green!40}0.962 \\
\hline
\textbf{Whole Task (Avg.)} & \cellcolor{green!40}\textbf{0.00} & \cellcolor{green!40}\textbf{0.10} & \cellcolor{green!40}\textbf{0.05} & \cellcolor{green!40}\textbf{0.960} \\
\hline
\end{tabular}
\end{table}

\subsection{Evaluation of Wrap Quality}

Beyond success rate, we propose the Paper Integrity Score (PIS) to quantify paper condition after manipulation:
\begin{equation} \label{Eq:1}
PI = 1 - [ w_1 T + w_2 W + w_3 E ]
\end{equation}
where $T$, $W$, and $E$ denote tear, wrinkle, and edge deformation scores. $w_1$, $w_2$, and $w_3$ are weight parameters.

A two-stage vision pipeline computes these scores: (i) the wrist-mounted camera captures a top-down image, segmented with SAM \cite{kirillov2023segment}, and (ii) a pre-trained YOLOv5 model \cite{luo2021you,li2024design} jointly regresses $T$, $W$, and $E$. 

Tears, wrinkles, and edge deformations differ visually and physically, making unified detection difficult. We adapted a pre-trained YOLOv5 model \cite{luo2021you,li2024design} to jointly regress the three defect scores, enabling holistic paper assessment.

Across 10 wrapping trials, results (Table \ref{tab:paper_integrity_results}) showed that lower defect scores aligned with higher PIS values. The overall average PIS was 0.962, indicating undamaged paper.





\begin{table*}[t]
\centering
\renewcommand{\arraystretch}{1.3}
\setlength{\tabcolsep}{6pt}
\caption{Ablation study results across framework modules.}
\resizebox{\linewidth}{!}{%
\begin{tabular}{|>{\columncolor[gray]{0.9}}l|l|l|l|l|}
\hline
\textbf{Ablated Module} & \textbf{Variant} & \textbf{Key Metric} & \textbf{Result vs. Full} & \textbf{Finding} \\ \hline

\textbf{Full Framework} & - & Task Success & \textbf{97\% (Baseline)} & - \\ \hline

\textbf{} & Zero-Shot & Plan Success (10 tasks) & 1\% (vs. 90\%) & Fine-tuning is critical \\ \cline{2-5}
\textbf{LLM Planner} & Fine-Tuned (Full) & Code Gen. Score (Avg. 1--4) & $\sim$4.0 (\textbf{Flawless}) & Output is a perfect API for Codex \\ \cline{2-5}
\textbf{} & Zero-Shot & Code Gen. Score (Avg. 1--4) & $\sim$1.9 (\textbf{Buggy}) & Natural language output is \textbf{not translatable} \\ \hline

\textbf{} & No Sub-Task ID & Success Rate & 35\% & Loses long-horizon context \\ \cline{2-5}
\textbf{Mid-Level Actor} & No F/T Data & Mean Force & +115\% & Uses force for adaptation \\ \cline{2-5}
\textbf{} & Manual Transition & Success Rate & 67\% & Smooth transitions needed \\ \hline

\textbf{Low-Level Control} & Position Only & Success Rate & 90\% (vs. 97\%) & Force compliance essential for precision \\ \hline

\end{tabular}
}
\label{tab:overall-results}
\end{table*}

\section{Ablation Studies} \label{Ablation}
To validate the design choices of our proposed framework, we conduct a series of ablation studies. Our goal is to quantify the contribution of each key component: the fine-tuned LLM high-level planner, the use of force feedback and admittance control, and the provision of a task identifier to the low-level policy. We made these studies to answer three questions:
\begin{enumerate}
    \item \textbf{Is the fine-tuned LLM necessary for understanding and sequencing tasks, or does a simpler method work just as well?}
    \item \textbf{ How crucial are the force conditioning and residual control for precise, contact-rich tasks?} 
    \item \textbf{Does the model rely on sub-task ID, or can it infer it from the scene?}
\end{enumerate}
\subsection{Ablating the High-Level Planner}
The LLM planner is responsible for translating natural language commands into structured plans. We ablate this module to test the hypothesis that task-specific fine-tuning is necessary for generating logically correct and executable code.
\textbf{Setup:} We compared our fine-tuned planner to a zero-shot baseline (GPT-4o, GPT-3.5) on 10 tasks (7 novel). Plans were evaluated on completeness, logical correctness, and executable specificity.
   \begin{figure}[t]
      \centering
      \includegraphics[width=0.5\textwidth]{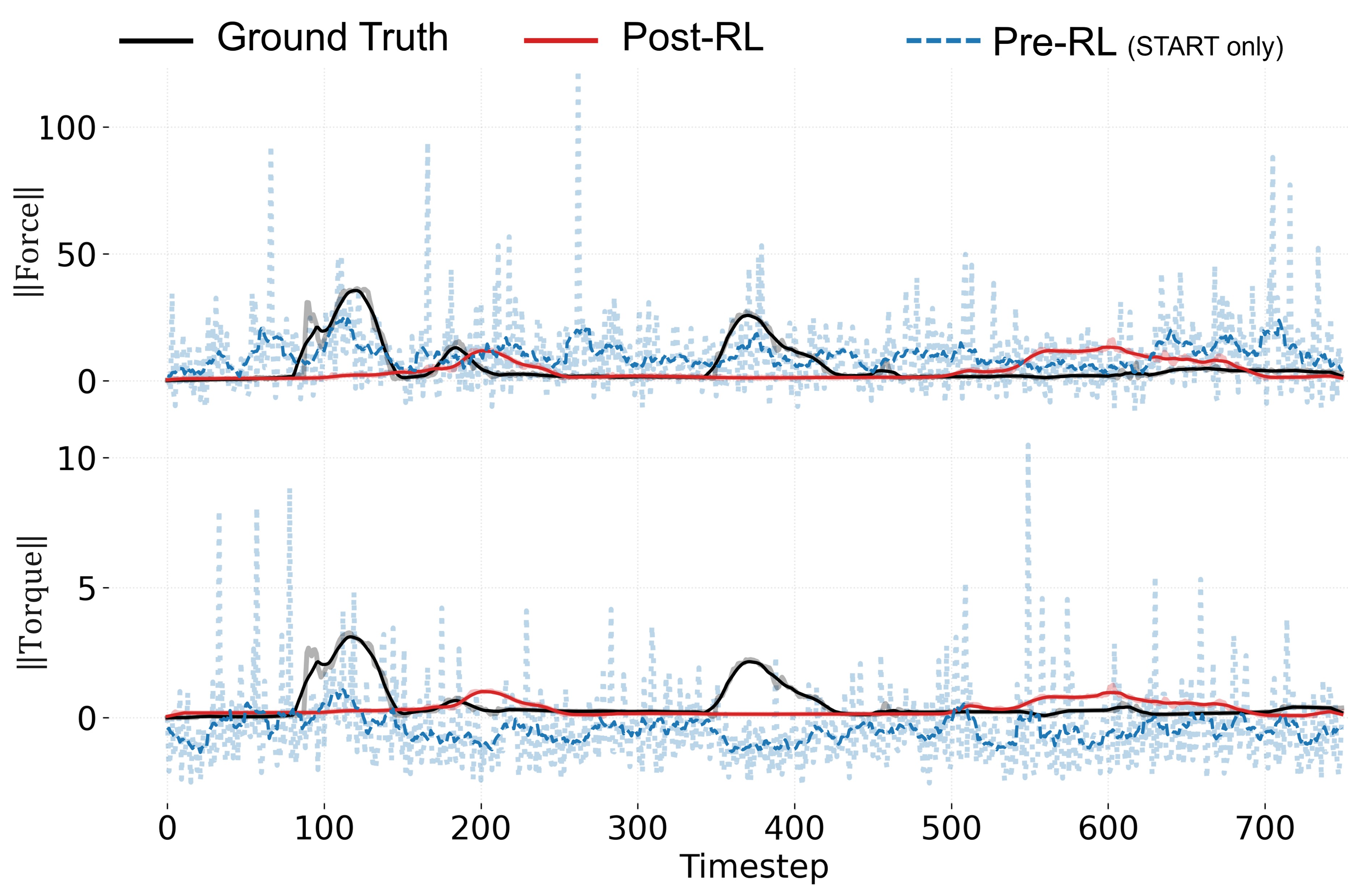}
      \caption{Comparison of force profiles during creasing. The RL-refined action closely tracks the ideal ground truth, unlike the raw START output, which overshoots or is too low, demonstrating precise and safe force control.}
      \label{fig:9}
   \end{figure}
\textbf{Results:}
\begin{itemize}
    \item \textbf{Sub-task ID:} The fine-tuned model showed a strong performance trend (80\% exact match vs. 20\% for zero-shot).
    \item \textbf{Full Plan Generation:} The fine-tuned planner achieved a \textbf{90\% success rate}, producing executable primitive sequences. The zero-shot planner failed on \textbf{all tasks (1\%)}, generating plans with critical logical errors (e.g., conflating human and robot roles) and ambiguity.
    \item \textbf{Code Translation:} The fine-tuned planner's output enabled reliable code generation (\textbf{$\sim$90\% predicted success}). The zero-shot's output was not translatable (\textbf{1\% predicted success}).
\end{itemize}

 \textbf{Implication: }Our fine-tuned planner is essential: it produces logically correct plans while formatting them for reliable translation into executable code. In contrast, the zero-shot approach fails at this step, highlighting the necessity of each component in the framework.
\vspace{0.4 cm}
\subsection{Ablating the Mid-Level Actor (START Model)}
\textbf{Experimental Setup:} We ablate key modifications to our START model to evaluate their contribution to long-horizon task execution. Specifically, we test: (1) removing sub-task ID conditioning, (2) removing force-torque (F/T) sensor input, and (3) replacing our learned transition flag with a manual heuristic.

\textbf{Results:} Preliminary results indicate that each modification is critical for robust execution. Removing the sub-task ID causes the policy to lose track of the overall task goal, reducing success rates on long-horizon tasks to 35\%, as the policy gets stuck in hovering at early stages. Removing F/T data from the policy input leads to a 115\% increase in average force exerted by the robot, indicating the policy uses force for high-level adaptation, not just low-level compliance. Finally, replacing the learned transition flag results in jerky transitions between sub-tasks and a 30\% drop in success rate.

\subsection{Ablating the Low-Level Corrector (RL-Based Admittance Controller)}
We evaluated the effect of the RL-based low-level corrector by comparing two conditions: (i) raw START actions executed with a stiff position controller, and (ii) START actions refined with the RL-tuned admittance controller. Raw actions from the START model often produced forces that were either too high, risking paper tearing, or too low, causing incomplete creases. In contrast, the RL-refined controller maintained forces within a safe and effective range, enabling stable creasing and preserving paper integrity (Fig.~\ref{fig:9}).

Removing the RL-based admittance controller severely degraded performance: forces became unstable, with frequent overshoot and oscillations, leading to tearing or failure to crease. Quantitatively, task success dropped from 97\% to 90\%, while average PIS decreased from 0.96 to 0.89. These results confirm that the RL module is essential for achieving compliant and adaptive interaction in contact-rich wrapping and folding tasks.

The overall results of the ablation studies are summarized in Table \ref{tab:overall-results}.
\subsection{Unified vs. Modular Policy}\label{uvsm}
The final experiment was done to examine whether our unified policy truly contributes to long-horizon robustness by comparing it against a modular baseline. In the modular setting, the wrapping sequence was segmented into five semantic subtasks, and a separate policy was trained for each. A learned phase switcher was used to activate the corresponding sub-policy during execution. Both approaches were trained with the same total number of demonstrations and network capacity to ensure fairness.

\textbf{Results:} Table \ref{tab:unified_vs_modular} shows the unified policy achieves higher end-to-end success. Modular policies perform well per sub-task but suffer from errors at boundaries, highlighting the advantage of a unified approach for maintaining global context.

\begin{table}[t]
\centering
\caption{Comparison of unified and modular policies.}
\label{tab:unified_vs_modular}
\begin{tabular}{lccc}
\toprule
Method   & Success Rate (\%) & Mean Time (s) & Peak Force (N) \\
\midrule
Unified  & \textbf{97}       & 170                  & 16.132 \\
Modular  & 90                & 175                  & 18.552 \\
\bottomrule
\end{tabular}
\end{table}
\section{Discussion} \label{Discussion}

Our experiments show that the trained policy generalizes robustly across environmental variations. Despite challenges such as task difficulty, logistical constraints, and limited visual contrast, the policy handled variations in paper stiffness, modest changes in box size, and different lighting conditions, with the visual pipeline effectively generalizing to light-colored, textured, or printed papers.

To manage wrapping complexity, the sequence is decomposed into $N_s = n$ semantic subtasks $\{\tau_1, \dots, \tau_n\}$, but unlike modular skill chaining, we learn a \textbf{single unified policy} $\pi_\theta: o_{1:T} \mapsto a_{1:T}$ across the full trajectory. This enables long-horizon generalization, dynamic adaptation to different task phases, and diverse actions for manipulating deformable objects such as paper, as validated in Section~\ref{uvsm}. The model also outputs a transition flag, allowing implicit switching between motion modes without external segmentation, which improves robustness and reduces reliance on hand-engineered task boundaries.

Our hybrid control architecture—combining START's long-horizon planning with local residual RL-based compliance—further enhances execution, particularly under material variability or unpredictable paper behavior.

A key limitation is failure on dark-colored papers, likely due to under-representation in training. Future work could use color augmentation to improve visual robustness and learn color-invariant features. Additionally, since the dataset contains only a single initial configuration, the policy does not generalize to different starting arrangements; geometric augmentation could address this limitation.

\section{Conclusion} \label{Conclusion}
This research presents a framework for robots to autonomously and collaboratively perform complex, long-horizon manipulation tasks with deformable objects, demonstrated in gift-wrapping. By combining sub-task aware imitation learning, residual reinforcement learning for force adaptation, and LLM-based task planning, the system robustly executes folding, creasing, and alignment, adapts to material variations, and supports human-robot collaboration. Results show that multi-modal demonstration learning with adaptive control improves task success in challenging scenarios. Future work could enhance generalization via data augmentation or video demonstrations, integrate tactile feedback, and extend planning with real-time perception and interactive dialogue.

\bibliography{refs} 



\end{document}